\documentclass[journal]{IEEEtran}
\usepackage{graphicx}
\usepackage{epstopdf}
\usepackage{dcolumn}% Align table columns on decimal point
\usepackage{bm}% bold math
\usepackage{float}
\usepackage{color}
\usepackage{subfigure}
\usepackage[utf8]{inputenc}
\usepackage[T1]{fontenc}
\usepackage{mathptmx}
\usepackage{booktabs} %需要加载宏包{booktabs}

\usepackage{algorithm}
\usepackage{algorithmic}

\ifCLASSINFOpdf
\else
\usepackage[dvips]{graphicx}
\graphicspath{{ /Users/lumioustang/Desktop/latex/Figure }}
\DeclareGraphicsExtensions{Fig2.eps}
\fi
\usepackage{amsmath}
\usepackage{amsmath} 
\usepackage{dcolumn}
\usepackage{algorithmic}
\usepackage{graphicx}
\usepackage{epstopdf}
\usepackage{dcolumn}% Align table columns on decimal point
\usepackage{bm}% bold math
\usepackage{float}
\usepackage{color}
\usepackage{subfigure}
\usepackage[utf8]{inputenc}
\usepackage[T1]{fontenc}
\usepackage{mathptmx}
\usepackage{tabularx, booktabs}
\usepackage[caption=false,font=normalsize,labelfont=sf,textfont=sf]{subfig}
\begin{document}
	%
	% paper title
	% Titles are generally capitalized except for words such as a, an, and, as,
	% at, but, by, for, in, nor, of, on, or, the, to and up, which are usually
	% not capitalized unless they are the first or last word of the title.
	% Linebreaks \\ can be used within to get better formatting as desired.
	% Do not put math or special symbols in the title.
	\title{{\color{black}Layer-wise Training Convolutional Neural Networks with Smaller Filters for Human Activity Recognition Using Wearable Sensors}}

	% author names and affiliations
	% transmag papers use the long conference author name format.
	
	\author{Yin~Tang,
		Qi~Teng,
		Lei~Zhang, Fuhong Min and Jun~He,~\textit{Member,~IEEE}
		\thanks{This work was supported  in part by the National Science Foundation of China under Grant 61971228 and the Industry-Academia Cooperation Innovation Fund Projection of Jiangsu Province under Grant BY2016001-02, and in part by the Natural Science Foundation of Jiangsu Province under grant BK20191371.\textit{ (Corresponding author: Lei Zhang.)}}% 
		\thanks{Yin Tang, Qi Teng, Lei Zhang and Fuhong Min are with School of Electrical and Automation Engineering, Nanjing Normal University, Nanjing, 210023, China (e-mail: leizhang@njnu.edu.cn). }% 
		
		\thanks{Jun He is with School of Electronic and Information Engineering, Nanjing
			University of Information Science and Technology, Nanjing 210044, China.}% 
	}

	% The paper headers
	%\markboth{Journal of \LaTeX\ Class Files,~Vol.~14, No.~8, August~2015}%
	%{Shell \MakeLowercase{\textit{et al.}}: Bare Demo of IEEEtran.cls for IEEE Transactions on Magnetics Journals}
	% The only time the second header will appear is for the odd numbered pages
	% after the title page when using the twoside option.
	% 
	% *** Note that you probably will NOT want to include the author's ***
	% *** name in the headers of peer review papers.                   ***
	% You can use \ifCLASSOPTIONpeerreview for conditional compilation here if
	% you desire.

	% If you want to put a publisher's ID mark on the page you can do it like
	% this:
	%\IEEEpubid{0000--0000/00\$00.00~\copyright~2015 IEEE}
	% Remember, if you use this you must call \IEEEpubidadjcol in the second
	% column for its text to clear the IEEEpubid mark.

	% use for special paper notices
	%\IEEEspecialpapernotice{(Invited Paper)}

	%\usepackage[mathlines]{lineno}% Enable numbering of text and display math
	%\linenumbers\relax % Commence numbering lines

	% for Transactions on Magnetics papers, we must declare the abstract and
	% index terms PRIOR to the title within the \IEEEtitleabstractindextext
	% IEEEtran command as these need to go into the title area created by
	% \maketitle.
	% As a general rule, do not put math, special symbols or citations
	% in the abstract or keywords.
	\IEEEtitleabstractindextext{%
		{\color{black}\begin{abstract}
			Recently, convolutional neural networks (CNNs) have set latest state-of-the-art on various human activity recognition (HAR) datasets. However, deep CNNs often require more computing resources, which limits their applications in embedded HAR. Although many successful methods have been proposed to reduce memory and FLOPs of CNNs, they often involve special network architectures designed for visual tasks, which are not suitable for deep HAR tasks with time series sensor signals, due to remarkable discrepancy. Therefore, it is necessary to develop lightweight deep models to perform HAR. As filter is the basic unit in constructing CNNs, it deserves further research whether re-designing smaller filters is applicable for deep HAR. In the paper, inspired by the idea, we proposed a lightweight CNN using Lego filters for HAR. A set of lower-dimensional filters is used as Lego bricks to be stacked for conventional filters, which does not rely on any special network structure. The local loss function is used to train model. To our knowledge, this is the first paper that proposes lightweight CNN for HAR in ubiquitous and wearable computing arena. The experiment results on five public HAR datasets, UCI-HAR dataset, OPPORTUNITY dataset, UNIMIB-SHAR dataset, PAMAP2 dataset, and WISDM dataset collected from either smartphones or multiple sensor nodes, indicate that our novel Lego CNN with local loss can greatly reduce memory and computation cost over CNN, while achieving higher accuracy. That is to say, the proposed model is smaller, faster and more accurate. Finally, we evaluate the actual performance on an Android smartphone.\\
		\end{abstract}}
		
		% Note that keywords are not normally used for peerreview papers.
		}

	% make the title area
	\maketitle

	% To allow for easy dual compilation without having to reenter the
	% abstract/keywords data, the \IEEEtitleabstractindextext text will
	% not be used in maketitle, but will appear (i.e., to be "transported")
	% here as \IEEEdisplaynontitleabstractindextext when the compsoc 
	% or transmag modes are not selected <OR> if conference mode is selected 
	% - because all conference papers position the abstract like regular
	% papers do.
	\IEEEdisplaynontitleabstractindextext
	% \IEEEdisplaynontitleabstractindextext has no effect when using
	% compsoc or transmag under a non-conference mode.

	% For peer review papers, you can put extra information on the cover
	% page as needed:
	% \ifCLASSOPTIONpeerreview
	% \begin{center} \bfseries EDICS Category: 3-BBND \end{center}
	% \fi
	%
	% For peerreview papers, this IEEEtran command inserts a page break and
	% creates the second title. It will be ignored for other modes.
	\IEEEpeerreviewmaketitle

	\section{Introduction}
	% The very first letter is a 2 line initial drop letter followed
	% by the rest of the first word in caps.
	% 
	% form to use if the first word consists of a single letter:
	% \IEEEPARstart{A}{demo} file is ....
	% 
	% form to use if you need the single drop letter followed by
	% normal text (unknown if ever used by the IEEE):
	% \IEEEPARstart{A}{}demo file is ....
	% 
	% Some journals put the first two words in caps:
	% \IEEEPARstart{T}{his demo} file is ....
	% 
	% Here we have the typical use of a "T" for an initial drop letter
	% and "HIS" in caps to complete the first word.
	%\IEEEPARstart{T}{his} demo file is intended to serve as a ``starter file''
	%for IEEE \textsc{Transactions on Magnetics} journal papers produced under \LaTeX\ using
	%IEEEtran.cls version 1.8b and later.
	% You must have at least 2 lines in the paragraph with the drop letter
	% (should never be an issue)
	%I wish you the best of success.

	\IEEEPARstart{w}{ith} the continuous technological advancement of mobile devices with sensing capabilities, ubiquitous sensing with the purpose of extracting knowledge from the data acquired by pervasive sensors, has become a very active research area. In particular, human activity recognition (HAR) using inertial sensors such as accelerometer and gyroscope embedded in smartphones or other edge devices has received much attention in recent years, due to the rapid growth of demand for various real-world applications such as smart homes, health monitoring, and sports tracking \cite{wang2019deep}. HAR can be considered as a typical pattern recognition (PR) problem, and traditional machine learning approaches such as decision tree \cite{casale2011human}, support vector machine \cite{anguita2012human} and naive Bayes \cite{zhang2004optimality} have made great achievement on inferring activity kinds. However, those conventional PR approaches may heavily rely on hand-crafted feature extraction \cite{berchtold2010actiserv}, which requires expert experience or domain knowledge. In the recent years, convolutional neural networks (CNNs) \cite{zeng2014convolutional}\cite{nweke2018deep}, represents the biggest trend in the field of machine learning, which can substitute for manually designed feature extraction procedures. Due to the emergence of CNN, research on machine learning is undergoing a transition from feature engineering to network engineering. Human efforts are shifting to designing smaller network architectures while keeping model performance. For a variety of HAR tasks, it has been widely demonstrated that building deeper CNN may result in higher performance {\color{black}\cite{qin2020imaging}}, but lead to the need for more resources such as memory and computational power. Deep models usually have millions of parameters, and their implementation on mobile devices becomes infeasible due to limited resources, which inevitably prevents the wide use of deep learning for HAR on mobile and wearable devices. Therefore, it is necessary to develop lightweight CNN to perform HAR.   \\
	\indent Recently, there has been rising interest in building small and efficient CNN for various embedded applications, whose goal is to reduce parameters while keeping model performance as much as possible. In particular, research in computer vision has been at the forefront of this work. This motivates a series of works towards lightweight network design, which can be generally categorized into either compressing pre-trained networks or designing small networks directly. For model compression {\color{black}\cite{cheng2018model}\cite{dieleman2016exploiting}\cite{ravanbakhsh2017equivariance}\cite{zhang2018learning}}, the existing works mainly focus on pruning, decomposing, parameters sharing or low-bit representing a basic network architecture, which cannot directly learn CNN from scratch. Due to the loss caused by compression, the performance of compressed model is usually upper bounded by its original pre-trained networks. These approaches often require special architectures and operation such as sparse convolution and fixed-point multiplication, which cannot be directly applied for HAR on off-the-shelf platform and hardware. An alternative is to design lightweight network architecture directly. For example, VGGNets \cite{simonyan2014very} and ResNets \cite{he2016deep} exhibit a simple yet efficient strategy of constructing deep networks: stacking building blocks of the same shape. Some researchers have demonstrated that carefully designed topologies are able to achieve compelling accuracy with low computational complexity. In particular, an important common idea is split-transform-merge \cite{xie2017aggregated}, in which the input is split into a few lower-dimensional embeddings, transformed by a set of specialized filters, and merged by concatenation. Based on the idea, Xception \cite{chollet2017xception}, MobileNet \cite{howard2017mobilenets}, Shufflenet \cite{zhang2018shufflenet} and ResNeXt \cite{sharma2018spatial} have achieved the state-of-the-art performance. However, the aforementioned approaches have seldom been directly adopted for HAR, according to related literatures.\\
	\indent The last few years have seen the success of network engineering in motion vision tasks as mentioned above, but it is still unclear how to adapt these architectures to new HAR dataset tasks, especially when there are remarkable different factors to be considered. In essence, HAR using inertial sensors can be seen as a classic multivariate time series classification problem, which makes use of sliding window to segment sensor signals and extracts discriminative features from them to be able to recognize activities by utilizing a classifier. Therefore, unlike imagery data, the HAR task has its own challenges. Though lightweight network modules achieve remarkable results in computer vision tasks, it has seldom been exploited in the field of HAR. As filter is a basic unit of constructing CNN, several researches have been conducted to discover whether it is applicable to re-design smaller filters in deep learning. Beyond the high-level network modules, Yang et al \cite{yang2019legonet} recently proposed an efficient CNN with Lego filters, which achieved state-of-the-art performance on motion vision tasks. For sensor based HAR, replacing ordinary filters with small Lego filters could be one feasible step to develop lightweight CNN deployed on mobile and wearable devices.\\
	\indent In this paper, we propose a lightweight CNN for HAR using Lego filters. To the best of our knowledge, building resource constrained deep networks suitable for HAR has never been explored, and this paper is the first try to develop lightweight CNN for HAR on ubiquitous and wearable computing area. Compared with standard convolution, convolution kernels constructed by lower dimensional Lego filters can greatly reduce the number of parameters. The Lego filters can be combined with the state-of-the-art deep models widely used in HAR, which enables substantially improved efficiency for various HAR applications. A method named as straight-through-estimator (STE) \cite{hubara2016binarized} is used to learn optimal permutation of Lego filters for a filter module in an end-to-end manner. A classic split-transform-merge three-stage strategy \cite{chollet2017xception}\cite{howard2017mobilenets} is utilized to further accelerate Lego convolutions. {\color{black}In our previous work \cite{teng2020layer}, layer-wise loss functions are used to train standard CNN. Without loss of generality, we train the Lego CNN with local loss, which can further improve performance without any extra cost.}\\
	\indent {\color{black}Deep models have powerful learning abilities, while shallow models are more efficient. To our knowledge, many model compression approaches have been proposed in computer vision field. How to perform both accurate and light-weight HAR still needs to be addressed. The design of lightweight CNN for HAR has been poorly explored in the literature. Without loss of generality, compression ratio and speedup are used to evaluate the performance of the proposed method.} The performance is evaluated on five public benchmark datasets, namely UCI-HAR dataset \cite{anguita2012human}, PAMAP2 dataset \cite{reiss2012introducing}, UNIMIB-SHAR dataset \cite{micucci2017unimib}, OPPORTUNITY dataset \cite{chavarriaga2013opportunity}, and WISDM dataset \cite{kwapisz2011activity}. {\color{black}Actually, it is expensive or even not affordable to collect enough “ground truth labeled” training data as benchmark in the realistic configuration of HAR.  To demonstrate the generality and superiority of the proposed method, we try to evaluate the performance across multiple most cited public HAR datasets, which are devised to benchmark various HAR algorithms. All the datasets are collected from either smartphones or multiple sensor nodes in ubiquitous and wearable computing scenarios. In particular, the authors of the OPPORTUNITY dataset have stated that the activity recognition environment and scenario has been designed to generate many activity primitives, yet in a realistic manner \cite{chavarriaga2013opportunity}. In the paper, our main research motivation is to develop a lightweight CNN for mobile and wearable computing. Therefore, we also evaluate the actual inference speed on a smartphone with an Android platform, which is cheaper and easier to use.} By comparing with the state-of-the-art methods on classification accuracy, memory and {\color{black}floating points operations per second} (FLOPs), we show how varying compression ratio affects over-all performance, and how such a lightweight system outperforms the state-of-the-art algorithms. The experiment results indicate the advantage of the lightweight CNN using Lego filters with regards to typical challenges for HAR in ubiquitous and wearable computing scenarios. {\color{black}Our main contribution is three-fold:\\
	\indent Firstly, {\color{black}in sensor based HAR scenarios we for the first time develop a lightweight CNN with smaller Lego filters}, which is able to greatly reduce memory and computation cost meanwhile {\color{black}maintaining almost the same accuracy;}\\
	\indent Secondly, we propose to train the Lego CNN with layer-wise loss functions, which can further improve results without any extra cost;\\
	\indent Thirdly, the experiment results indicate that the proposed method can consistently outperform the baseline CNN on test error. When compared to our previous method with local loss {\color{black}\cite{teng2020layer}}, the layer-wise training Lego CNN can achieve almost the same state-of-the-art performance, even though the number of parameters and FLOPs are much smaller. That is to say, the proposed method is smaller, faster and more accurate.}\\
	\indent The paper is structured as follows. Section II summarizes related works of HAR and deep compression. Section III presents the details of deep local loss HAR using Lego filters. Section IV details the HAR dataset, experimental setup used, and our experimental results. In Section V, we extend and discuss above experiment results and in Section VI, we draw our conclusions. 
	\section{related works}
	\indent In recent years, due to advances of the computational capabilities, CNN have achieved remarkable results on sensor based HAR \cite{janidarmian2017comprehensive} and outperformed other state-of-the-art algorithms which requires advanced preprocessing or cumbersome hand-crafting feature extraction. For example, Zeng et al \cite{zeng2014convolutional} firstly applied CNN to HAR, which extracts the local dependency and scale invariant characteristics of the acceleration time series. Yang et al \cite{yang2015deep} applied CNN with hierarchical models to demonstrate its superiority to traditional shallow machine learning methods on several benchmark HAR datasets. Jiang et al \cite{jiang2015human} transformed the raw sensor signal into 2D image signal, and then a two layer CNN is used to classify this signal image equaling to the desired activity recognition. Hammerla et al \cite{hammerla2016deep} did an early work by evaluating the performance of various deep learning techniques through 4000 experiments on some public HAR datasets. Teng et al \cite{teng2020layer} proposed a layer-wise CNN using local loss function, which can achieve state-of-the-art performance across multiple benchmark HAR datasets. Wang et al \cite{wang2019attention} proposed an attention based CNN to perform weakly labeled HAR tasks, which can greatly facilitate the process of sensor data annotation. Ordonez et al \cite{ordonez2016deep} proposed a new DeepConvLSTM architecture composed of CNN and recurrent networks, which outperforms CNN. {\color{black}Agarwal et al \cite{agarwal2020lightweight} proposed a lightweight Recurrent Neural Network (RNN) in HAR applications.} On the whole, shallow neural networks and conventional PR methods could not achieve good performance, compared with deep learning. However, deep models often require lots of computing resources, which is not available for HAR using mobile and wearable devices \cite{wang2019deep}\cite{wang2020sequential}. Thus it deserves deep research into lightweight CNN architecture of better performance for HAR.\\
	\indent Recent research effort on visual recognition has been shifting to design small network with high performance. In particular, when there are more layers, designing network architectures becomes increasingly difficult due to the growing number of hyper-parameters. The increasing demands for running efficient deep neural networks on embedded devices also encourage the study. Several representative state-of-the-art networks are reviewed. SqueezeNet \cite{iandola2016squeezenet} in early 2016 was the first paper that was concerned with building a memory efficient architecture. VGGNets \cite{simonyan2014very} tend to reduce free choices of hyper-parameters by stacking building block of same shape to construct network, and this strategy is also inherited by ResNets \cite{he2016deep}. Another important strategy is split-transform-merge as mentioned above. Based on this strategy, Google’s MobileNets \cite{howard2017mobilenets} goes one step further by modifying the standard convolutional operation as depth-wise separable convolution and point-wise convolution. The idea of depth-wise convolutions in MobileNets is then generalized to group-wise convolutions as in ShuffleNets \cite{zhang2018shufflenet}. Designing convolution with a compact filter can effectively reduce the computation cost. The key idea is to replace the loose and over-parametric filters with compact blocks to improve network performance. As filter is the basic unit in CNNs, Yang et al \cite{yang2019legonet} recently used re-designed Lego filters to accelerate convolutions, which achieved state-of-the-art performance. Despite the success of deep compression on computer vision, the primary use of the aforementioned models mainly lies in image or video tasks, which have seldom been directly adopted to perform HAR. In the next section, we will describe the convolution operation constructed by Lego filters, and then present the entire architecture of the lightweight CNN used in HAR.
	\section{model}
	\indent In this section, the lightweight CNN architecture with Lego filters termed as Lego CNN is proposed to handle the unique challenges existed in HAR. The challenges in HAR \cite{wang2019deep} problem usually include (i) processing units (i.e., filters) in CNN need applied along temporal dimension and (ii) sharing the units in CNN among multiple sensors. For HAR, we deal with multiple channels of time series signals, in which the traditional CNN cannot be used directly. The sliding window strategy is adopted to segment the time series signals into a collection of short pieces of signals. Hence the signals are split into windows of a fixed size and an overlap between adjacent windows is tolerated for preserving the continuity of activities. An instance handled by CNN is a two-dimensional matrix with r raw samples, in which each sample contains multiple sensor attributes observed at time t. Here, r is the number of samples per window. For comparison, the baseline model is built as a typical deep CNN, which comprises of convolutional layers, dense layers and softmax layers. Our research aims to realize lightweight CNN for the practical use of HAR. Following the settings of Yang et al \cite{yang2019legonet}, the ordinary filters are replaced with a set of compact Lego filters, that are often of much lower dimensions. As filter is the basic unit in CNN, the ordinary convolution filters stacked by a set of lower-dimensional Lego filters can lead to an efficient model. Instead of manually stacking these Lego filters, we realize convolution operation by simultaneously optimizing Lego filters and their combination (i.e., binary masks) at the training stage of deep neural networks. For binary masks, gradient-based learning is infeasible. Alternatively, the Straight-Through-Estimator (STE) is used in the discrete optimization problem with gradient descent due to its effectiveness and simplicity. As these filter modules share the same set of Lego filters but with different combinations, without loss of generality, a classical split-transform-merge three-stage strategy is adopted to further accelerate convolutions by exploiting intermediate feature maps. An overview of the proposed lightweight HAR system is shown in Fig. 1.
	\begin{figure}[htbp]
		\hspace*{0cm}
		\centering
		\includegraphics[scale=0.45]{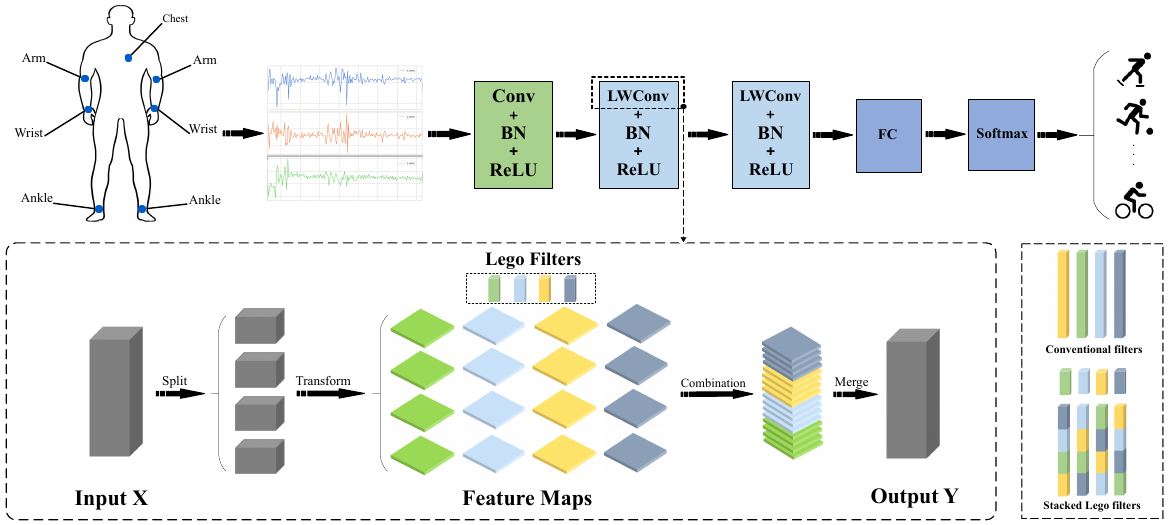}% Here is how to import EPS art
		%\graphicspath{{'G:\360MoveData\Users\Administrator\Desktop\LATEX_LCOAL_ERROR'}}
		%\DeclareGraphicsExtensions{Fig1.eps}
		\caption{\label{Fig1}Overview of the model framework with Lego CNN for HAR. {\color{black}This figure shows how the three-stage pipeline split-transform-merge operates on input feature maps. \textbf{X} is the input feature map, and \textbf{Lego Filters} are convolved with different segmented fragments from \textbf{X}, which result in a set of \textbf{Feature Maps}. \textbf{Y} is generated by merging the intermediate feature maps.}}
	\end{figure}
	\subsection{Lego Filters for Constructing CNNs}
	\indent As mentioned above, CNN has achieved the state-of-the-art performance in HAR. Without loss of generality, a common convolutional layer with n filters can be represented as $F=\left\{ f_1,\ f_2,\ ..., \ f_n \right\} \in R^{d\times 1\times c\times n}$, where $d\times 1$ is the size of filters and c is the channel number. The conventional convolution operation can be represented as: $Y=X^TF$, where X and Y are the input and output feature maps of this layer. The filters F can be solved by using the standard feed-forward and back-propagation method. {\color{black}As shown in the bottom right corner of Fig. 1, F is replaced with a set of smaller filters $B=\left\{ b_1,\ b_2,\ ..., \ b_k \right\} \in R^{d\times 1\times \tilde{c}\times k}$ with fewer channels ($\tilde{c}\ll c$), namely Lego filters \cite{yang2019legonet}, which can be represented as: F = BM, where M is a linear transformation of stacking Lego filters. F is used as a filter module, as it is assembled with Lego filters. Each Lego filter can be utilized for multiple times in constructing a filters module F. Hence, convolutional filters constructed by these Lego filters B of fewer parameters can be solved from the following optimization problem:}\\
	\begin{equation}
	\hat{B}=arg\underset{B}{\min}\frac{1}{2}\lVert Y,\ L\left( BM  ,\ X \right) \rVert _{2}^{F}
	\end{equation}
where $
\lVert \cdot \rVert _F
$ is the Frobenius norm for matrices.
	\subsection{Combining Lego Filters and Optimization}
	For the use of Lego filters, the X is split into o=c/$\tilde{c}$ fragments $\left[ X_1,\ ..., \ X_o \right]$, and k Lego filters are stacked for a matrix $B=\left[ vec\left( b_1 \right) ,\ ...,\ vec\left( b_k \right) \right] \in R^{d\times 1\times \tilde{c}\times k}$. Note that each output feature map is the sum of convolutions on all fragments of the input feature maps. e.g., The j-th feature map $Y^{j}$ formed by the j-th Lego convolutional filter can be formulated as:
	\begin{equation}
	Y^j=\sum_{i=1}^o{X_{i}^{T}\left( BM_{i}^{j} \right)}
	\end{equation}
	where $M_{i}^{j}\in \left\{ 0,1 \right\} ^{k\times 1}$ and $\lVert M_{i}^{j} \rVert _1=1$ is a binary mask. As there is the constraint on M with $\lVert M_{i}^{j} \rVert _1=1$, only one Lego filter can be selected from B for the i-th fragment of the input feature maps, which ensure that Lego filters can be concatenated brick by brick. Therefore, the above optimization problem for simultaneously learning Lego filters and their combination in Eq. 1 can be rewritten as:
	\begin{equation}
	\begin{array}{l}
	\ \ \ \ \ \ \ \underset{B,M^j}{\min}\sum_{i=1}^o{\frac{1}{2}\lVert Y^j-X_{i}^{T}\left( BM_{i}^{j} \right) \rVert _{F}^{2}} \\
	s.t.\ M_{i}^{j}\in \left\{ 0,1 \right\} ^{k\times 1},\ \lVert M_{i}^{j} \rVert _1=1,i=1,...,o\\ 
	\end{array}
	\end{equation}
	
	Evidently, M is a binary matrix which is difficult to optimize using Adam. To solve the optimization problem, the object function can be relaxed by introducing $N \in R^{n\times o\times k}$ whose shape is equivalent to M. For model training, M can be binarized from N as follows:
	\begin{equation}
	\begin{array}{l}
	M_{i,k}^{j}=\begin{cases}
	1,&		if\ k=arg\max N_{i}^{j}\\
	0,&		otherwise\\
\end{cases}\\
        \ \ \ s.t.\ j=1,...,n,\ i=1,...,o
        \end{array}
	\end{equation}
	\ \ \ The gradient $\varDelta N$ for float parameters N is equivalent to the gradient $\varDelta M$ . The STE is used for back-propagating gradients throughout the quantitation function \cite{yang2019legonet}\cite{hubara2016binarized}.
	\subsection{{\color{black}More Efficient Convolution}}
	In the previous procedure, convolution filters are firstly constructed by a set of Lego filters, and then applied on input feature maps. As these filter modules share the same set of Lego filters but with different combinations, repeated computations will be introduced during the convolution stage. A classical split-transform-merge strategy \cite{howard2017mobilenets}\cite{yang2019legonet} is used to remove these repeated computations and further accelerate convolutions. This split-transform-merge pipeline is introduced as follows:\\
	$\textbf{1. Split: }$The X is split into o fragments $\left[ X_1,\ ..., \ X_o \right]$, in which each fragment $X_i$ will be the feature map with smaller channels to be convolved with s set of Lego filters.\\
        $\textbf{2. Transform: }$ The o fragments are convolved with each individual Lego filter, i.e., which leads to $o\times k$ intermediate feature maps in total. The convolution process can be represented as:	
	\begin{equation}
	I_{ij}=X_{i}^{T}B_j
	\end{equation}
	$\textbf{3. Merge: }$From the perspective of matrix, Eq. 2 can be easily rewritten as:
	\begin{equation}
	Y^j=\sum_{i=1}^o{\left( X_{i}^{T}B \right) M_{i}^{j}}
	\end{equation}
	where the $X_{i}^{T}B$ is the intermediate feature map I. To remove repeated computations and accelerate convolutions, M extracts intermediate feature maps from I and merge them to produce the output feature maps Y. 
	\subsection{{\color{black}Lego CNN with Local Loss}}
	{\color{black}In order to get better performance, we propose a new layer-wise training Lego CNN using local loss for sensor based HAR. For local loss functions, the computational graph is detached after each hidden layer to prevent standard backward gradient flow. {\color{black}Referring to previous work using layer-wise loss functions \cite{teng2020layer}}, the global loss in Lego CNN is replaced with two local loss functions \cite{teng2020layer}\cite{nokland2019training}. One of the local loss signal is implemented by the cross entropy between a prediction of local linear classifier and the target, which is called prediction loss $L_{pred\_loss}$. It can be expressed as follows:
	\begin{equation}
	L_{pred\_loss}=CrossEntropy\left( Y,\ W^TX \right) 
	\end{equation}
	where W denotes a linear classifier, X is the output of a forward-flow convolutional layer and Y denotes the label matrix of one-hot encoded targets.\\
	\indent The other loss function is similarity matching loss \cite{nokland2019training}, which is formulated as follows:
	\begin{equation}
	L_{sim\_loss}=\lVert S\left( C\left( X;w \right) \right) -S\left( Y \right)\rVert _2 \centering
	\end{equation}
	where C represents a convolutional operation with kernel size 3*3, stride 1 and padding 1. The S(*) denotes the adjusted cosine similarity matrix operation.\\
	\indent Finally, the weighted combination of the above loss functions can be represented as:
	\begin{equation}
	L_{local\_loss}=\left( 1-\alpha \right) L_{pred\_loss}+\alpha L_{sim\_loss}
	\end{equation}
	in which $\alpha$ is a weighting factor and is set to 0.99 according to our previous work \cite{teng2020layer}.\\}
	\section{experiment}
	\indent The experiments are conducted on five public datasets including UCI-HAR dataset, OPPORTUNITY dataset, UNIMIB-SHAR dataset, PAMAP2 dataset and WISDM dataset, which is typical for HAR in ubicomp (described below). The CNN composed of several convolutional layers and one fully connected layer was used as the baseline to evaluate whether the Lego filters can reduce the number of parameters while keeping performance. {\color{black}Actually, the baseline CNN structure is commonly used, which is constituted by (i) a convolution layer that convolves the input or the previous layer’s output with a set of kernels to be learned; (ii) a rectified linear unit (ReLU) layer that maps the output of the previous layer by the function $relu\left( \nu \right) = \max \left( \nu ,0 \right)$; (iii) a normalization layer that normalizes the values of different feature maps in the previous layer. As it is hard to know all specific CNN structures used in other HAR literatures on five benchmark datasets, the baseline CNN is trained via tuning hyper-parameters, which achieve almost the same accuracy obtained in these HAR literatures \cite{hammerla2016deep}\cite{ignatov2018real}\cite{li2018comparison}. We conclude that the baseline CNN has comparable feature extracting and classification ability. The performance is compared between proposed CNN, baseline CNN and other state-of-the-art in the experiment part.} Batch normalization was applied before each ReLU activation function. Although there are lots of parameters in the last fully connected layer, the Lego filters are not used to compress the last layer in all our experiments. If Lego filters are used to compress the last layer, many classes would share similar features, which would inevitably introduce side effects and deteriorate the classification performance of HAR. The Lego filters are not applied in the first convolutional layer as the size of conventional filter is often small in this layer. That is to say, only the intermediate convolutional layers are compressed with Lego filters. \\
	\indent The different compression rates are explored throughout whole experiments. There are two parameters, e.g., o and m, used to tune compression ratio in Lego CNN. The o is an integer which indicates the number of fragments input feature maps are split into, and the m is a decimal smaller than one which indicates the ratio of Lego filters compared to the original of each layer, i.e., $\frac{k}{n}$. Here, it is evident that the number of Lego filters k should be smaller than the output channel number n. Since binary matrix M is much smaller than Lego filters parameters, the compression ratio for each convolutional layer can approximately be calculated as $\frac{n\times o}{k}$. Similarly, the theoretical speedup for an optimized convolution layer using smaller Lego filters can approximately be calculated as $\frac{n}{k}$. However, as mentioned above, the Lego filters have not been applied for each layer and the actual compression ratio cannot attain the aforementioned theoretical upper limit.\\ 
	\indent In a fully supervised way, the network parameters are optimized by minimizing the cross-entropy loss function with mini-batch gradient descent using an Adam optimizer. The network will be trained at least 500 epochs. The initial learning rate and batch size were set according to different datasets. Since no clear consensus exists on which sliding window size should be preferably employed for deep learning, for comparison, the same values used in previous case of success are selected. As human activity datasets are often highly unbalanced, the overall classification accuracy is not an appropriate measure to evaluate HAR tasks. Requiring performance metrics that are independent of the class distribution, we evaluate the models using the weighted F1 score \cite{ordonez2016deep}:
	\begin{equation}
	F_1=2\sum{\frac{N_c}{N_{total}}\frac{Pecision_c\times \text{Re}call_c}{Pecision_c+\text{Re}call_c}}
	\end{equation}
	which considers the correct classification of each class equally important. $N_c$ is the number of samples in class c, and $N_{total}$ is the total number of samples. The experiments are repeated 5 times and the mean F1 score is used as the final measure to evaluate model performance. The model training and classification are run in PyTorch (Paszke et al, 2017 \cite{paszke2017automatic}) deep learning framework on a machine with an Intel i7-6850K CPU, 32GB RAM and NVIDIA RTX 2080 Ti GPU.\\
	\subsubsection{$\textbf{The OPPORTUNITY dataset}$ {\color{black}\cite{chavarriaga2013opportunity}}}
	\indent The dataset contains a set of complex naturalistic activities collected in a sensor-rich environment, {\color{black}which is comprised of the readings of various motion sensors recorded:\\
	\indent$\bullet$ Body-worn sensors: 7 inertial measurement units, 12 3D acceleration sensors, 4 3D localization information;\\
	\indent$\bullet$ Object sensors: 12 objects with 3D acceleration and 2D rate of turn;\\
	\indent$\bullet$ Ambient sensors: 13 switches and 8 3D acceleration sensors.}\\
 \indent During the recordings, participants were asked to perform a session five times with activities of daily living (ADL) and one drill session. The dataset is publicly available and can be downloaded from the UCI Machine Learning repository, which has been used in an open activity recognition challenge. In this paper, we train and test our models on the same subset used in the OPPORTUNITY challenge, which is composed of the recordings of 4 subjects including only on-body sensors. Data is preprocessed at a frequency of 30Hz from 12 locations on the body, and annotated with 18 mid-level gesture annotations. \\
	\indent In the experiment, for each subject, data from 5 different ADLs is recorded. ADL1, ADL2 and ADL3 from subject 1, 2 and 3 is used as our training set via replicating the most popular recognition challenge with ADL4 and ADL5 from subject 4 and 5 in our test set. For frame-by-frame analysis, the sliding windows size is 64 and the sliding step is 8. The resulting training set contains approximately 650k samples. For the dataset, the shorthand description of the baseline CNN is C(128)$\rightarrow$C(256)$\rightarrow$C(384)$\rightarrow$FC$\rightarrow$$S_m$, where C($L^s$) denotes a convolutional layer with $L^s$ feature maps, FC a dense layer and $S_m$ a softmax classifier. The two intermediate convolutional layers with Lego filters are used. The batch size is set to 300 and learning rate was set constant to 5e-4.
	
	\begin{figure}[htbp]
		\hspace*{0cm}
		\centering
		\includegraphics[scale=0.1]{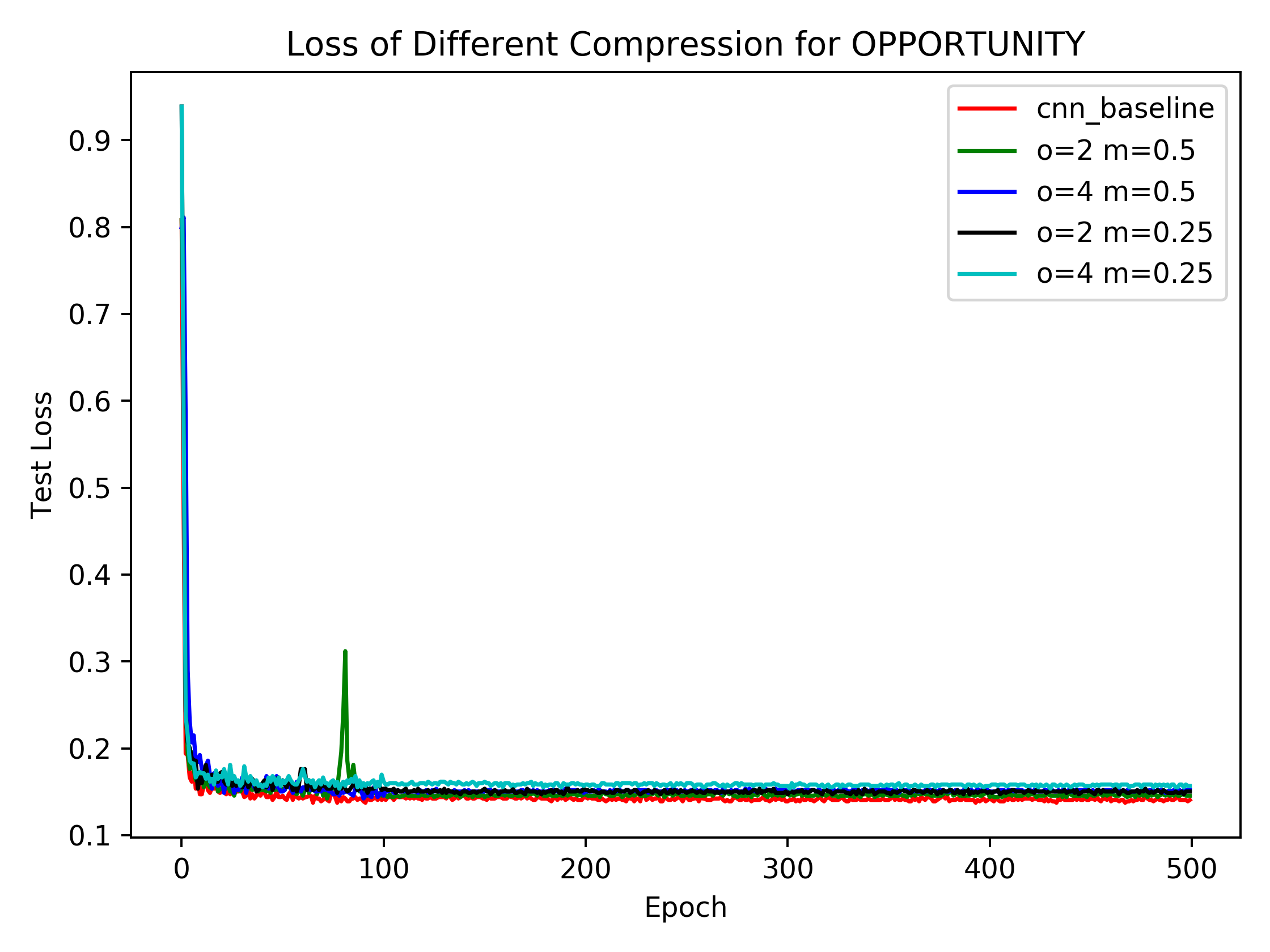}% Here is how to import EPS art
		%\graphicspath{{'G:\360MoveData\Users\Administrator\Desktop\LATEX_LCOAL_ERROR'}}
		%\DeclareGraphicsExtensions{Fig1.eps}
		\caption{\label{Fig2} Overview of the loss of different compression for \textbf{OPPORTUNITY}.}
	\end{figure}
	\indent As there is a notable imbalance in the OPPORTUNITY dataset where the NULL class represents 72.28$\%$, the model performance is evaluated considering the NULL class. Fig. 2 shows the effect of increasing compression ratio on the performance with Lego CNN(o=2,m=0.5), Lego CNN(o=4,m=0.5), Lego CNN(o=2,m=0.25), and Lego CNN (o=4,m=0.25) architectures. As mentioned above, different o and m is set to change compression ratio. These results of the baseline CNN approach those obtained previously by Yang et al, 2015 {\color{black}\cite{yang2015deep}} using a CNN on raw signal data. From the results in Fig. 2, it can be seen that the baseline CNN consistently outperforms Lego CNN, which agrees well with our motivation. Compared with the baseline CNN, there is no significant decrease in performance on test data with increasing compression ratio. Table I presents classification accuracy, memory and FLOPs for the different compression rates on the OPPORTUNITY dataset. {\color{black}It can be seen that the baseline CNN achieves 86.10$\%$ accuracy. }When compared to the best submissions using CNN for the OPPORTUNITY challenge, accuracy drops less than 1.6$\%$, e.g., Lego CNN(o=4,m=0.25). However, it can be noticed that Lego CNN offers a striking performance improvement: there is a 7.6x compression ratio and 3x speedup in terms of FLOPs. In other words, the Lego filters can efficiently compress networks without increasing any computational burden, which is suitable for HAR applications on mobile devices.
	
	\begin{table}[h]
		\caption{Performance of Different Compression for \textbf{OPPORTUNITY}}
		\centering
		\begin{tabular}{cccccc}
			\toprule 
			\textbf{Model}&{\color{black}\textbf{F1}}&\textbf{Memory}&\textbf{Com}&\textbf{FLOPs}&\textbf{Speed Up}\\
			\midrule
			Baseline&\textbf{86.10$\%$}&3.20M&1.0x&41.90M&1.0x\\
			\midrule
			o=2,m=0.5&86.01$\%$&0.95M&3.4x&23.15M&1.8x\\
			o=4,m=0.5&85.46$\%$&0.61M&5.3x&23.15M&1.8x\\
			o=2,m=0.25&85.48$\%$&0.61M&5.3x&13.78M&3.0x\\
			o=4,m=0.25&84.50$\%$&0.42M&7.6x&13.78M&3.0x\\
			\bottomrule
			\label{Tab1}
		\end{tabular}
	\end{table}

	\subsubsection{$\textbf{The PAMAP2 dataset}$ {\color{black}\cite{reiss2012introducing}}}	
	\indent The dataset consists of 18 different physical activities such as \emph{house cleaning, watching TV, rope jumping, playing soccer}, etc. As instructed, all subjects performed 12 different activities, and some of the subjects performed 6 optional activities. The collector aggregated data from 9 subjects wearing 3 inertial measurement units (IMUs) and a heart rate monitor, where the 3 IMUs were placed over the wrist, chest and ankle on the dominant. The heart rate is recorded at a sampling frequency of 9Hz. The IMUs are sampled at a frequency of 100Hz. \\
	\indent For comparison, the accelerometer signals are subsampled to 33.3Hz, which has been used in other HAR literatures. To generate a larger number of segments, we sliced the sensor data using sliding window size corresponds to 5.12 s, which allows a 78$\%$ overlapping rate. For the PAMAP2 dataset, we randomly split 80$\%$ for training and 20$\%$ for test. Considering that there are many categories of dataset, we increase the number of convolution layers and the shorthand description of the baseline CNN is C(128)$\rightarrow$C(256)$\rightarrow$C(384)$\rightarrow$C(512)$\rightarrow$C(512)$\rightarrow$FC$\rightarrow$$S_m$ including 5 convolutional layers and 1 fully connected layer. The batch size is set to 300 and the initial learning rate was set to 1e-4. The learning rate is reduced by a factor of 0.1 after 100 epochs.
	\begin{figure}[htbp]
		\hspace*{0cm}
		\centering
		\includegraphics[scale=0.1]{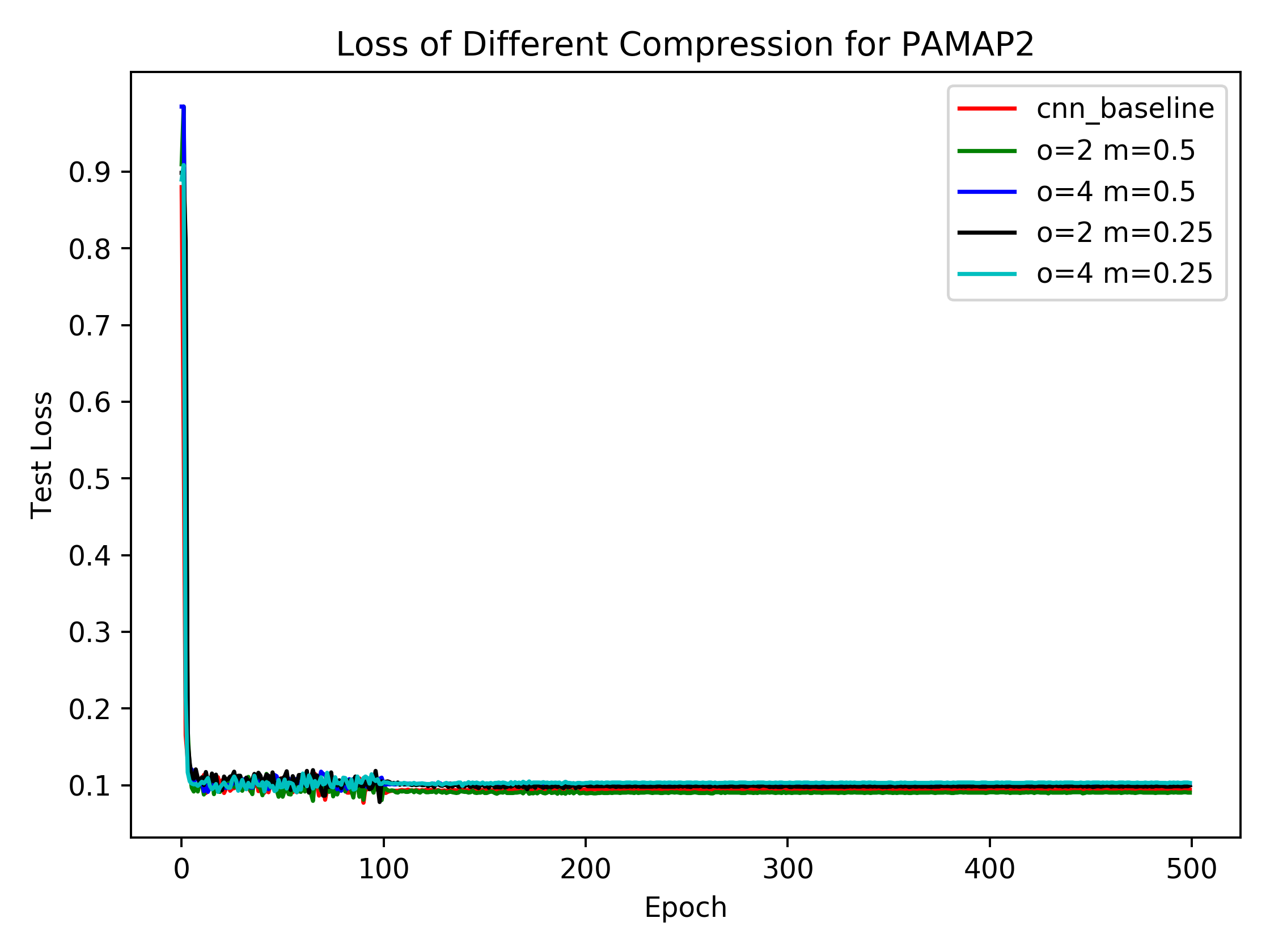}% Here is how to import EPS art
		%\graphicspath{{'G:\360MoveData\Users\Administrator\Desktop\LATEX_LCOAL_ERROR'}}
		%\DeclareGraphicsExtensions{Fig1.eps}
		\caption{\label{Fig3} Overview of the loss of different compression for \textbf{PAMAP2}.}
	\end{figure}

	\indent In Fig. 3, the performance of the different compression ratio is evaluated on the PAMAP2 dataset.The baseline CNN approach cannot offer better results than the Lego CNN(o=2,m=0.5). Table II shows the relationship between performance and two parameters. It can be seen that the baseline CNN achieves 91.26$\%$ accuracy, which approaches the previous reported results using a CNN (Yang et al, 2018 {\color{black}\cite{yang2018dfternet}}). The Lego CNN with a range of o and m systematically outperforms baseline in terms of memory and FLOPs. The Lego CNN(o=2,m=0.5) presents the best accuracy, which achieves 0.14$\%$ improvement on baseline. Note that the baseline network has about 2.8 times more parameters than the Lego CNN. As compression ratio increases, the performance of the model slightly decreases. When setting o=4 and m=0.25, we are able to achieve less than 1$\%$ accuracy drop with a compression ratio of 5x and a speedup of 3.9x. That is to say, without any extra cost, we can train a lightweight CNN using Lego filters with almost the same accuracy.\\
	\begin{table}[h]
		\caption{Performance of Different Compression for \textbf{PAMAP2}}
		\centering
		\begin{tabular}{cccccc}
			\toprule 
			\textbf{Model}&{\color{black}\textbf{F1}}&\textbf{Memory}&\textbf{Com}&\textbf{FLOPs}&\textbf{Speed Up}\\
			\midrule
			Baseline&91.26$\%$&7.93M&1.0x&305.16M&1.0x\\
			\midrule
			o=2,m=0.5&\textbf{91.40$\%$}&2.86M&2.8x&154.48M&2.0x\\
			o=4,m=0.5&90.91$\%$&2.02M&3.9x&154.48M&2.0x\\
			o=2,m=0.25&90.65$\%$&2.02M&3.9x&79.13M&3.9x\\
			o=4,m=0.25&90.52$\%$&1.60M&5.0x&79.13M&3.9x\\
			\bottomrule
			\label{Tab2}
		\end{tabular}
	\end{table}

	\subsubsection{$\textbf{The UCI-HAR dataset}$ {\color{black}\cite{anguita2012human}}}	
	\indent The UCI-HAR dataset has been collected from a group of 30 subjects within an age bracket of 19-48 years. Each subject, wearing a Samsung Galaxy S II smartphone on the waist, was asked to perform six activities including \emph{walking, walking\_upstairs, walking\_downstairs, sitting, standing and laying.} The three axial linear acceleration and three axial angular velocity were recorded at a constant sample rate of 50Hz by using the embedded accelerometer and gyroscope in the smartphone. The dataset has been labeled manually by video-recorded.\\
	\indent The accelerometer and gyroscope signals are pre-processed by applying noise filters. In particular, the accelerometer signals are composed of gravitational and body motion components, where the gravitational force is assumed to have only low frequency components. Therefore, the above two components were further separated by using a Butterworth low-pass filter with 0.3 Hz cutoff frequency. The sensor signals were then sampled by using a fixed-width sliding windows of 128 and 50$\%$ overlap (2.56s/window). For the experiment, the dataset has been randomly partitioned into two sets where 70$\%$ of the subjects was selected for generating training data and 30$\%$ for test data. For the UCI-HAR dataset, the shorthand description of the baseline CNN is C(128)$\rightarrow$C(256)$\rightarrow$C(384)$\rightarrow$FC$\rightarrow$$S_m$. The model was trained using Adam optimizer with mini-batch size of 200. The learning rate is reduced by a factor of 0.1 after 100 epochs, and the initial learning rate was set to 4e-4.
	\begin{figure}[htbp]
		\hspace*{0cm}
		\centering
		\includegraphics[scale=0.1]{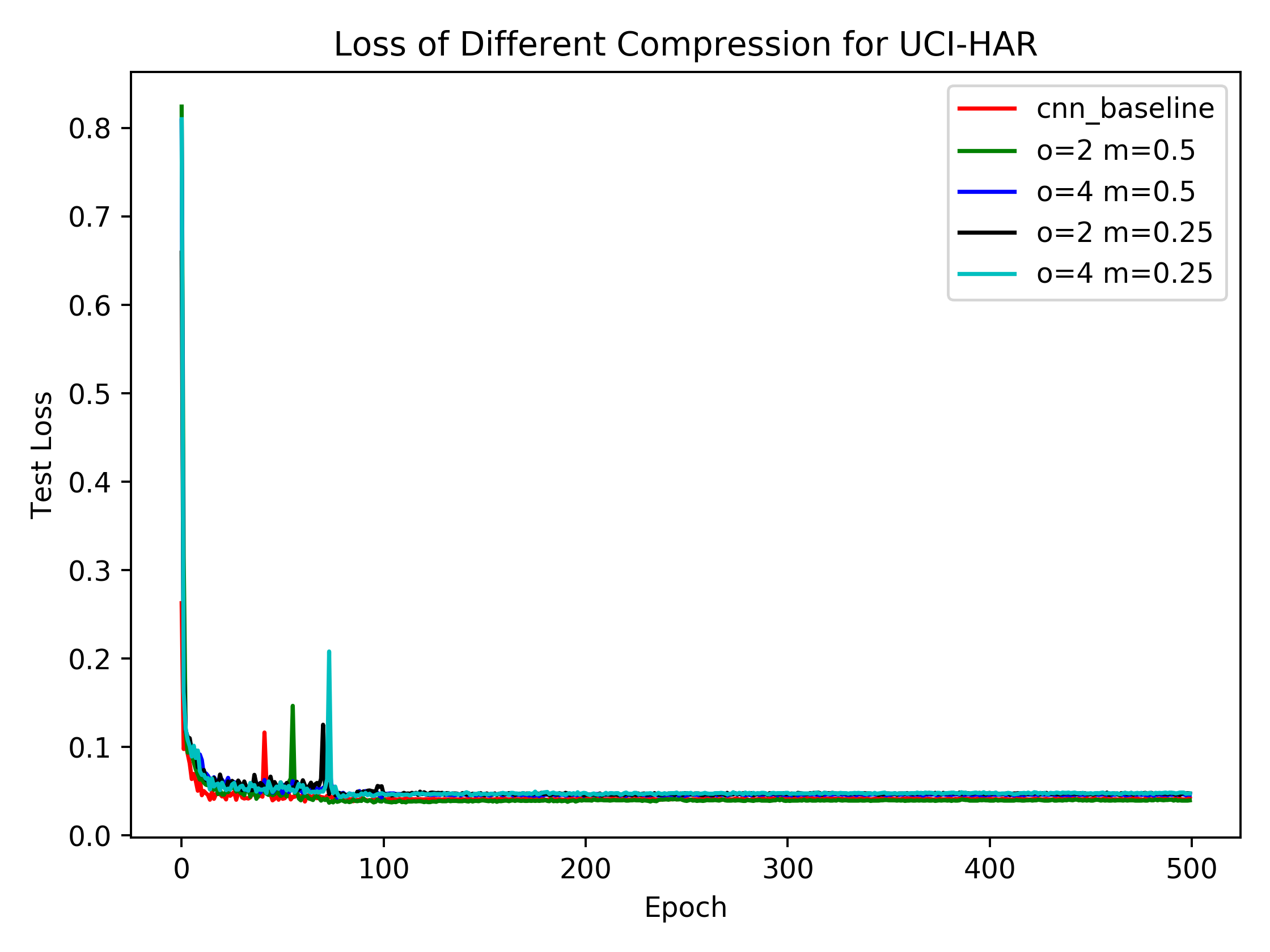}% Here is how to import EPS art
		%\graphicspath{{'G:\360MoveData\Users\Administrator\Desktop\LATEX_LCOAL_ERROR'}}
		%\DeclareGraphicsExtensions{Fig1.eps}
		\caption{\label{Fig4} Overview of the loss of different compression for \textbf{UCI-HAR}.}
	\end{figure}

	\indent Using the above experiment configurations, we then increase the compression ratio. Fig. 4 shows the effect of increasing compression ratio on performance. Compared with the baseline CNN, the Lego CNN(o=2,m=0.5) achieves higher performance on the test set with less parameters. The Lego CNN is compared with the state-of-the-art CNN based methods in HAR, as seen in the Table III. The best published results on this task to our knowledge is 97.62$\%$ using CNN combined with hand-crafted features(Ignatov et al, 2018 {\color{black}\cite{ignatov2018real}}). To make the comparison more fair, we train the baseline CNN without using other techniques, which achieves 96.23$\%$ accuracy, almost in line with the results using CNN alone by Jiang et al \cite{jiang2015human}. Comparison shows that the Lego CNN(o=2,m=0.5) even outperforms the baseline CNN, achieving an accuracy of 96.27$\%$ with a compression ratio of 2.7x and a speedup of 2x. We argue that if parameters are not too few, parameters are enough to learn comparable or even better results. There is a continuous decrease on performance as compression ratio increases. Accuracy only drops 0.73$\%$ than the baseline CNN accompanied by a compression ratio of 5.2x and a speedup of 3.9x , which is acceptable for the mobile HAR task.
	\begin{table}[h]
		\caption{Performance of Different Compression for \textbf{UCI-HAR}}
		\centering
		\begin{tabular}{cccccc}
			\toprule 
			\textbf{Model}&{\color{black}\textbf{F1}}&\textbf{Memory}&\textbf{Com}&\textbf{FLOPs}&\textbf{Speed Up}\\
			\midrule
			Baseline&96.23$\%$&3.55M&1.0x&69.62M&1.0x\\
			\midrule
			o=2,m=0.5&\textbf{96.27$\%$}&1.30M&2.7x&34.99M&2.0x\\
			o=4,m=0.5&95.90$\%$&0.92M&3.9x&34.99M&2.0x\\
			o=2,m=0.25&95.92$\%$&0.92M&3.9x&17.67M&3.9x\\
			o=4,m=0.25&95.50$\%$&0.69M&5.2x&17.67M&3.9x\\
			\bottomrule
			\label{Tab3}
		\end{tabular}
	\end{table}

	\subsubsection{$\textbf{The UNIMIB-SHAR dataset}$ {\color{black}\cite{micucci2017unimib}}}
	\indent This dataset is a new dataset which includes 11,771 samples for the use of HAR and fall detection. The dataset aggregates data from 30 subjects (6 male and 24 female whose ages ranging from 18 to 60 years) acquired using a Bosh BMA220 3D accelerometer of a Samsung Galaxy Nexus I9250 smartphone. The data are sampled at a frequency of 50 Hz, which is commonly used in the related literature for HAR. The whole dataset consists of 17 fine grained classes, which is further split into 9 types of ADLs and 8 types of falls. The dataset also stores related information used to select samples according to different criteria, such as the type of ADL performed, the gender, the age, and so on.\\
	\indent Unlike OPPORTUNITY, there is no any NULL class in the UNIMIB SHAR dataset, which remains fairly balanced. For this dataset, the sliding windows of data and their associated labels are directly produced with a fixed length T = 151, which corresponds to approximately 3s. The sliding step length is set to 3. The dataset contains 11,771 time windows of size 151*3 in total. In the experiment, the dataset is randomly divided into two parts where 70$\%$ was selected to generate training data and 30$\%$ test data. For the UNIMIB-SHAR dataset, the network structure of the baseline of CNN is C(128)$\rightarrow$C(256)$\rightarrow$C(384)$\rightarrow$FC$\rightarrow$$S_m$, which has 3 convolutional layers and 1 fully connected layer. The model was trained using Adam optimizer with mini-batch size of 200, and the learning rate was set to 5e-4.
	\begin{figure}[htbp]
		\hspace*{0cm}
		\centering
		\includegraphics[scale=0.1]{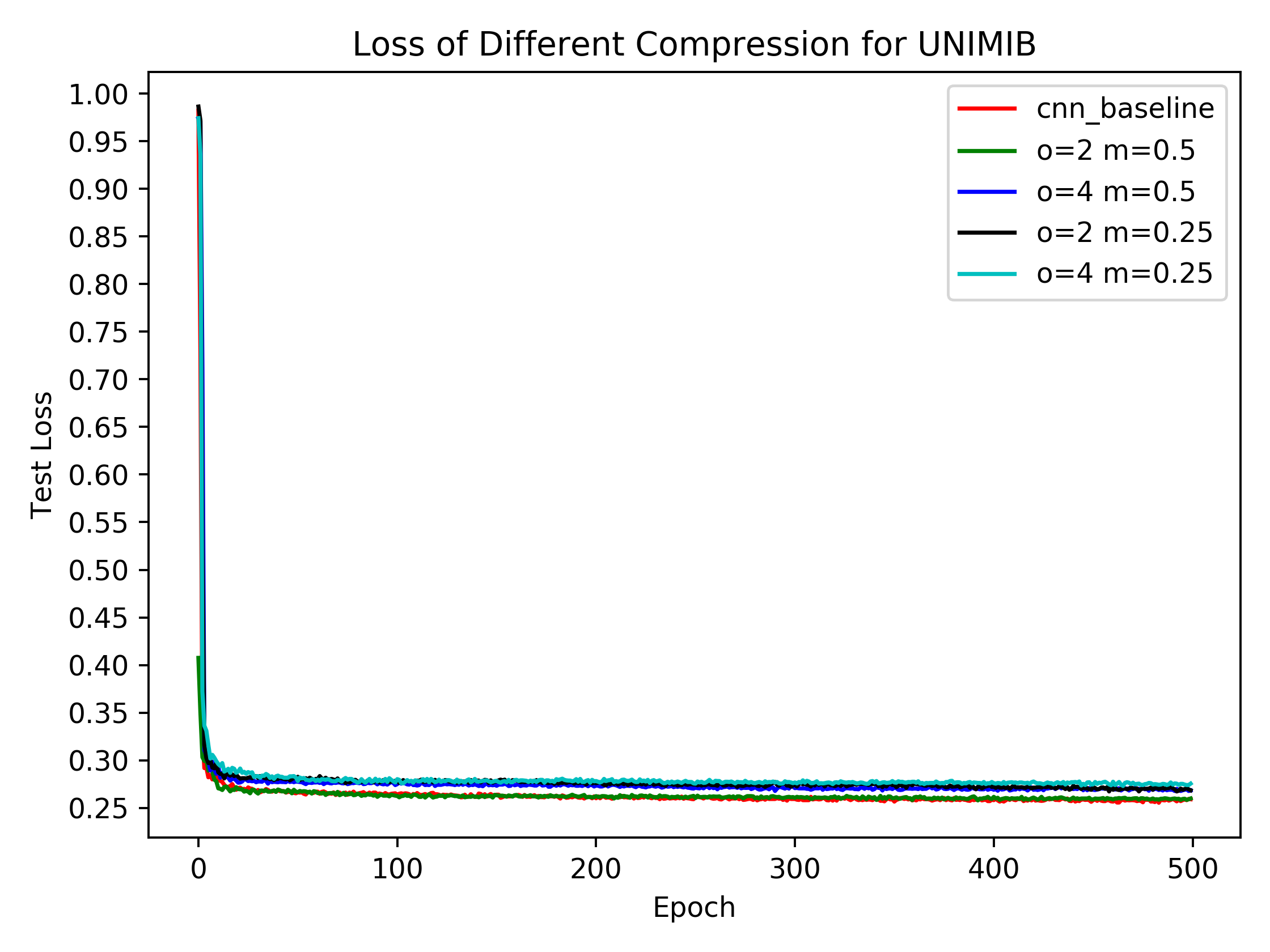}% Here is how to import EPS art
		%\graphicspath{{'G:\360MoveData\Users\Administrator\Desktop\LATEX_LCOAL_ERROR'}}
		%\DeclareGraphicsExtensions{Fig1.eps}
		\caption{\label{Fig5} Overview of the loss of different compression for \textbf{UNIMIB-SHAR}.}
	\end{figure}
	\\
	\indent Fig. 5 demonstrates the performance of Lego CNN with a range of compression ratio compared with the baseline CNN. The Lego CNN(o=2,m=0.5) achieves almost the same accuracy with the baseline CNN, and there is a steady slight decrease in performance on test data with increasing compression ratio. Table IV demonstrates the performance of our model compared with the state-of-the-arts using CNN in terms of accuracy, compression ratio, memory, and FLOPs. To our knowledge, the best result reported using CNN on this dataset is 74.66$\%$ (Li et al, 2018 {\color{black}\cite{li2018comparison}}), which is consistent with our results of CNN. It can be noticed that accuracy of Lego CNN(o=2,m=0.5) only drops 0.05$\%$ less than that of the baseline CNN. Accuracy keeps to be almost the same with 3.1x compression ratio. Meanwhile, FLOPs reduced a lot in the model by approximately 2x. In the extreme compression situation of 6.6x, the Lego CNN with coefficient o=4, m=0.25 still could maintain performance about 72.80$\%$ accuracy, compared to 74.46$\%$ accuracy of the baseline CNN. From results in the table, the Lego CNN is more portable alternative to the existing state-of-the-art HAR applications using CNN.
	
	\begin{table}[h]
		\caption{Performance of Different Compression for \textbf{UNIMIB-SHAR}}
		\centering
		\begin{tabular}{cccccc}
			\toprule 
			\textbf{Model}&{\color{black}\textbf{F1}}&\textbf{Memory}&\textbf{Com}&\textbf{FLOPs}&\textbf{Speed Up}\\
			\midrule
			Baseline&\textbf{74.46$\%$}&5.80M&1.0x&40.73M&1.0x\\
			\midrule
			o=2,m=0.5&74.41$\%$&1.87M&3.1x&20.47M&2.0x\\
			o=4,m=0.5&73.27$\%$&1.22M&4.8x&20.47M&2.0x\\
			o=2,m=0.25&73.25$\%$&1.22M&4.8x&10.33M&3.9x\\
			o=4,m=0.25&72.80$\%$&0.88M&6.6x&10.33M&3.9x\\
			\bottomrule
			\label{Tab4}
		\end{tabular}
	\end{table}
	
	\subsubsection{$\textbf{The WISDM dataset}$ {\color{black}\cite{kwapisz2011activity}}}
	\indent This WISDM dataset contains 1098213 samples which belong to 29 subjects. One triaxial accelerometer embedded in mobile phones with Android OS is used to generate data. In a supervised condition, the smartphones were placed in a front leg pocket of the dominant. Each subject performed 6 distinctive human activities of \emph{walking, jogging, walking upstairs, walking downstairs, sitting and standing}. The acceleration signals were recorded at a constant sampling rate of 20Hz (Kwapisz, Weiss, and Moore 2011 \cite{kwapisz2011activity}). In the experiment, the accelerometer signals were preprocessing by the sliding window technique. The sliding windows size was set to 10s and the sliding step length was set to 1s, which allows a 90$\%$ overlapping rate. The whole WISDM dataset was randomly split into two parts where 70$\%$ was selected to generate training data and the rest test data. The shorthand description of the baseline CNN is C(64)$\rightarrow$C(128)$\rightarrow$C(256)$\rightarrow$C(256)$\rightarrow$C(384)$\rightarrow$FC$\rightarrow$$S_m$, which has 5 convolutional layers and 1 fully connected layer. The network will be trained with the batch size of 200 using the conventional Adam optimizer. The initial learning rate is set as 0.001, which will be reduced by a factor of 0.1 after each 100 epochs.
	\begin{figure}[htbp]
		\hspace*{0cm}
		\centering
		\includegraphics[scale=0.1]{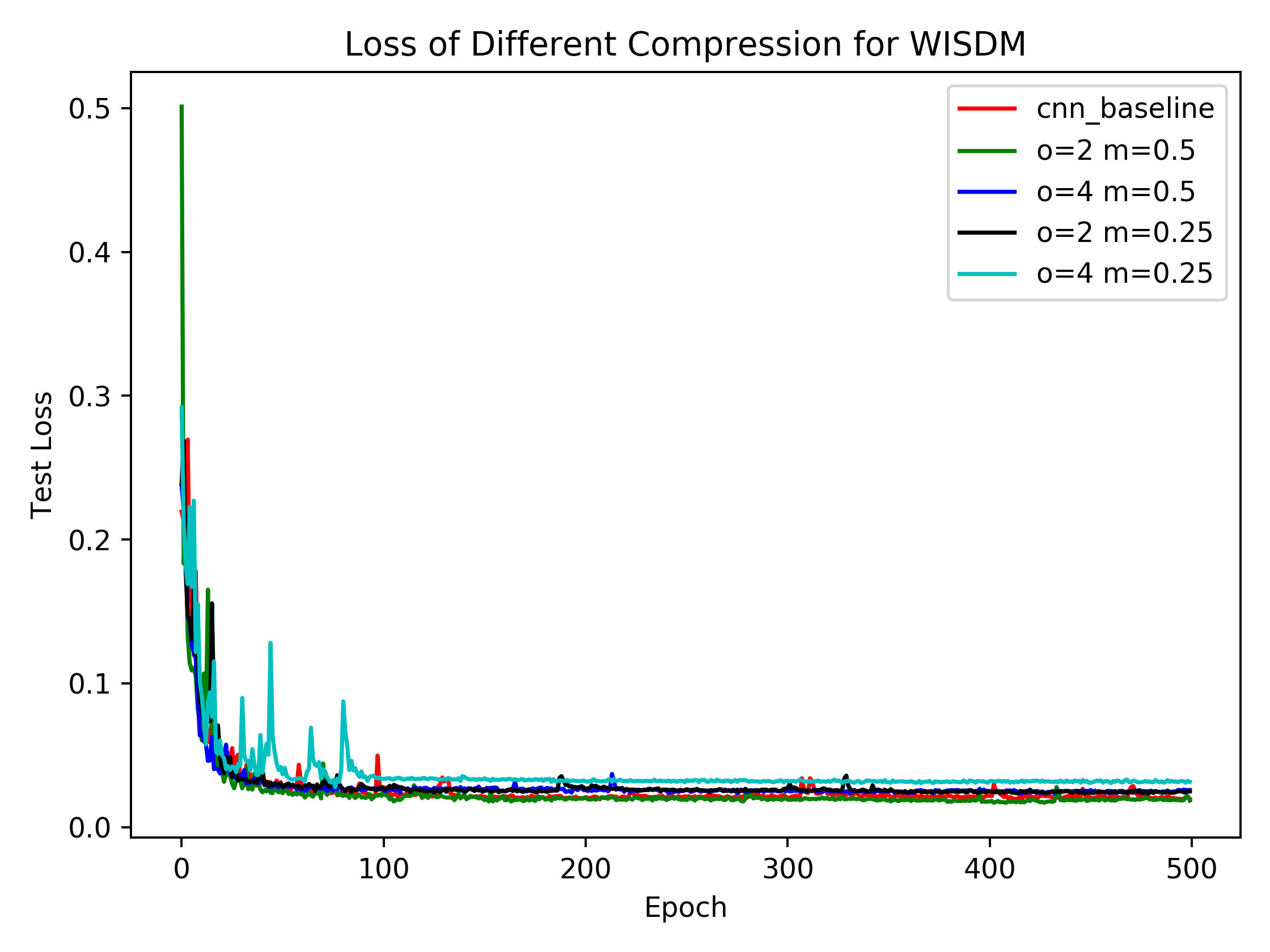}% Here is how to import EPS art
		%\graphicspath{{'G:\360MoveData\Users\Administrator\Desktop\LATEX_LCOAL_ERROR'}}
		%\DeclareGraphicsExtensions{Fig1.eps}
		\caption{\label{Fig6} Overview of the loss of different compression for \textbf{WISDM}.}
	\end{figure}\\
	\indent As seen in Fig. 6, there is no significant decrease on performance of the Lego CNN with moderate compression rates. According to the test curve, the Lego CNN(o=2,m=0.5) even is able to achieve higher performance than baseline. The performance of Lego CNN is compared with the baseline CNN on the WISDM dataset. To our knowledge, the best published results using a CNN on this dataset is 98.2$\%$ using spectrogram signals instead of raw acceleration data (Ravi et al, 2016 {\color{black}\cite{ravi2016deep}}; Alsheikh et al, 2015 {\color{black}\cite{alsheikh2016deep}}). Since main research motivation in the paper is to discuss lightweight CNN using Lego filters, for simplicity, we still train the baseline CNN with raw acceleration data, which achieve 97.30$\%$ accuracy, slightly lower than above results. From results in the Table V, what you can see is that the Lego CNN(o=2,m=0.5) achieves 0.21$\%$ performance improvement on baseline with a compression ratio of 3.1x and a speedup of 2x. And it is worth mentioning that, even in the extreme situation, the Lego CNN(o=4,m=0.25) achieves only 1$\%$ accuracy drop than baseline with a compression ratio of 6.8x and a speedup of 4x.
	\begin{table}[h]
		\caption{Performance of Different Compression for \textbf{WISDM}}
		\centering
		\begin{tabular}{cccccc}
			\toprule 
			\textbf{Model}&{\color{black}\textbf{F1}}&\textbf{Memory}&\textbf{Com}&\textbf{FLOPs}&\textbf{Speed Up}\\
			\midrule
			Baseline&97.30$\%$&5.15M&1.0x&132.09M&1.0x\\
			\midrule
			o=2,m=0.5&\textbf{97.51$\%$}&1.64M&3.1x&66.11M&2.0x\\
			o=4,m=0.5&96.90$\%$&1.07M&4.8x&66.11M&2.0x\\
			o=2,m=0.25&96.92$\%$&1.07M&4.8x&33.12M&4.0x\\
			o=4,m=0.25&96.30$\%$&0.76M&6.8x&33.12M&4.0x\\
			\bottomrule
			\label{Tab5}
		\end{tabular}
	\end{table}
	\begin{figure}[htbp]
		\hspace*{0cm}
		\centering
		\includegraphics[scale=0.105]{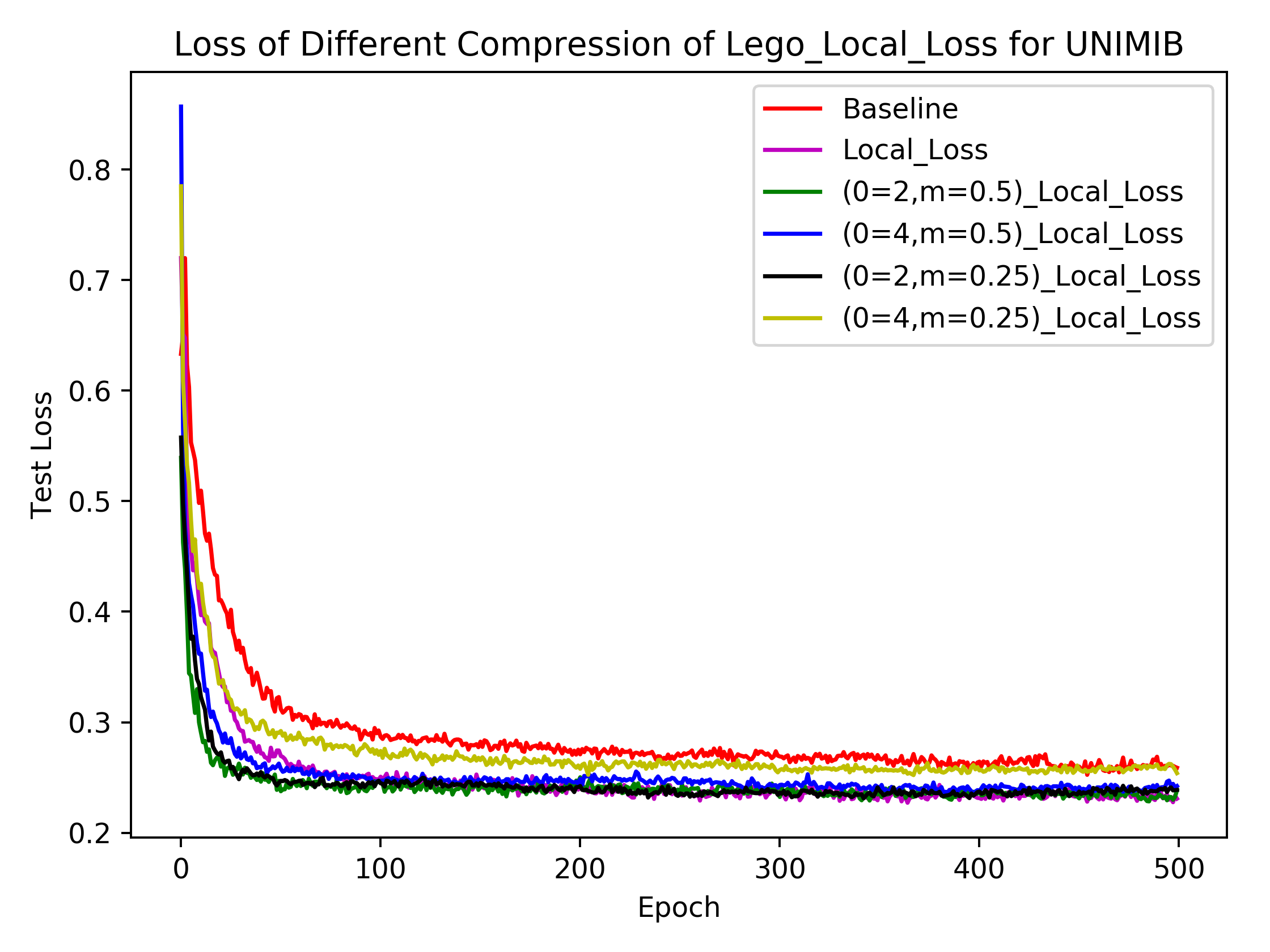}% Here is how to import EPS art
		%\graphicspath{{'G:\360MoveData\Users\Administrator\Desktop\LATEX_LCOAL_ERROR'}}
		%\DeclareGraphicsExtensions{Fig1.eps}
		\caption{\label{Fig7} {\color{black}Overview of the loss of Lego-Local Loss for \textbf{UNIMIB-SHAR}.}}
	\end{figure}
	\subsubsection{$\textbf{Lego CNN with Local Loss}$}
	\indent {\color{black}Fig.7 illustrates that the layer-wise training Lego CNN (under different o, m) is able to consistently outperform baseline for test error on the UNIMIB-SHAR dataset. As stated in our previous article \cite{teng2020layer}, the local loss can play a regulating effect, which leads to final higher training errors. Thus, the proposed method may achieve lower test errors, due to a better generalization ability. When compared to our previous results using local loss, there is no significant decrease on accuracy, even though the number of parameters and FLOPs was much smaller (Table VI).} Under different compression ratio,  the layer-wise training Lego CNN with local loss is also compared with baseline as well as local loss method, on several other benchmark datasets. As can be seen in Table VII, the results imply that the new model is smaller, faster and more accurate.
	\vspace{0cm} 
	\begin{table}[h]
	\newcommand{\tabincell}[2]{\begin{tabular}{@{}#1@{}}#2\end{tabular}}
		\caption{{\color{black}Performance with  Lego-Local Loss for UNIMIB-SHAR}}
		\centering
		\begin{tabular}{cccc}
			\toprule 
			\textbf{Model}&{\color{black}\textbf{F1}}&\textbf{Memory}&\textbf{FLOPs}\\
			\bottomrule
			Baseline&\textbf{74.46$\%$}&-&-\\
			\bottomrule
			Local$\_$Loss&\textbf{77.80$\%$}&-&-\\
			\midrule
			o=2,m=0.5&77.50$\%$&0.62M&20.47M\\
		
			o=4,m=0.5&76.23$\%$&0.41M&20.47M\\
		
			o=2,m=0.25&76.10$\%$&0.41M&10.33M\\
		
			o=4,m=0.25&75.01$\%$&0.29M&10.33M\\
			\bottomrule
			\label{Tab6}
		\end{tabular}
	\end{table}
	\vspace{0cm} 
	\begin{table}[h]
	\newcommand{\tabincell}[2]{\begin{tabular}{@{}#1@{}}#2\end{tabular}}
		\caption{{\color{black}Accuracy with  Lego-Local Loss for different datasets}}
		\centering
		\begin{tabular}{ccccc}
			\toprule 
			\textbf{Model}&\textbf{UCI-HAR}&\textbf{PAMAP2}&\textbf{WISDM}&\textbf{OPPORTUNITY}\\
			\bottomrule
			Baseline&96.23$\%$&91.26$\%$&97.30$\%$&86.10$\%$\\
			\bottomrule
			Local$\_$Loss&\textbf{96.90$\%$}&92.97$\%$&\textbf{98.82$\%$}&\textbf{88.09$\%$}\\
			\midrule
			o=2,m=0.5&96.80$\%$&\textbf{93.50$\%$}&98.80$\%$&87.90$\%$\\
		
			o=4,m=0.5&96.55$\%$&92.38$\%$&98.08$\%$&87.01$\%$\\
		
			o=2,m=0.25&96.60$\%$&92.25$\%$&98.11$\%$&86.87$\%$\\
		
			o=4,m=0.25&96.32$\%$&91.50$\%$&97.60$\%$&86.55$\%$\\
			\bottomrule
			\label{Tab6}
		\end{tabular}
	\end{table}
	\\
	\\

		\vspace{-1cm}  
	\section{discussion}
	\indent {\color{black} Throughout the whole experiments, there are two tunable parameters in Lego CNN, in which o indicates the number of fragments input feature maps are split into and m indicates the ratio of Lego filters compared to the original of each layer, i.e., $\frac{k}{n}$. Different compression ratios can be achieved by setting o or m. As compression ratio grows, accuracy often drops, which will lead to a decrease on memory or FLOPs. Thus there is a trade-off between accuracy, memory and FLOPs. Actually, memory can directly indicates the final compression performance of the model. Under the same memory budget, the experiment result show that higher o with much more fragments still achieve almost the same accuracy, but takes much more FLOPs. For example, the Lego CNN(o=4,m=0.5) can achieve comparable accuracy with the Lego CNN(o=2,m=0.25), but costs almost twice FLOPs. With the same memory, one should choose smaller o to balance memory and FLOPs, which is also in line with Yang et al’s \cite{yang2019legonet} results on visual tasks.\\}
	\begin{figure}[htbp]

	\begin{minipage}{0.5\linewidth}
		{\centering\includegraphics[width=9cm]{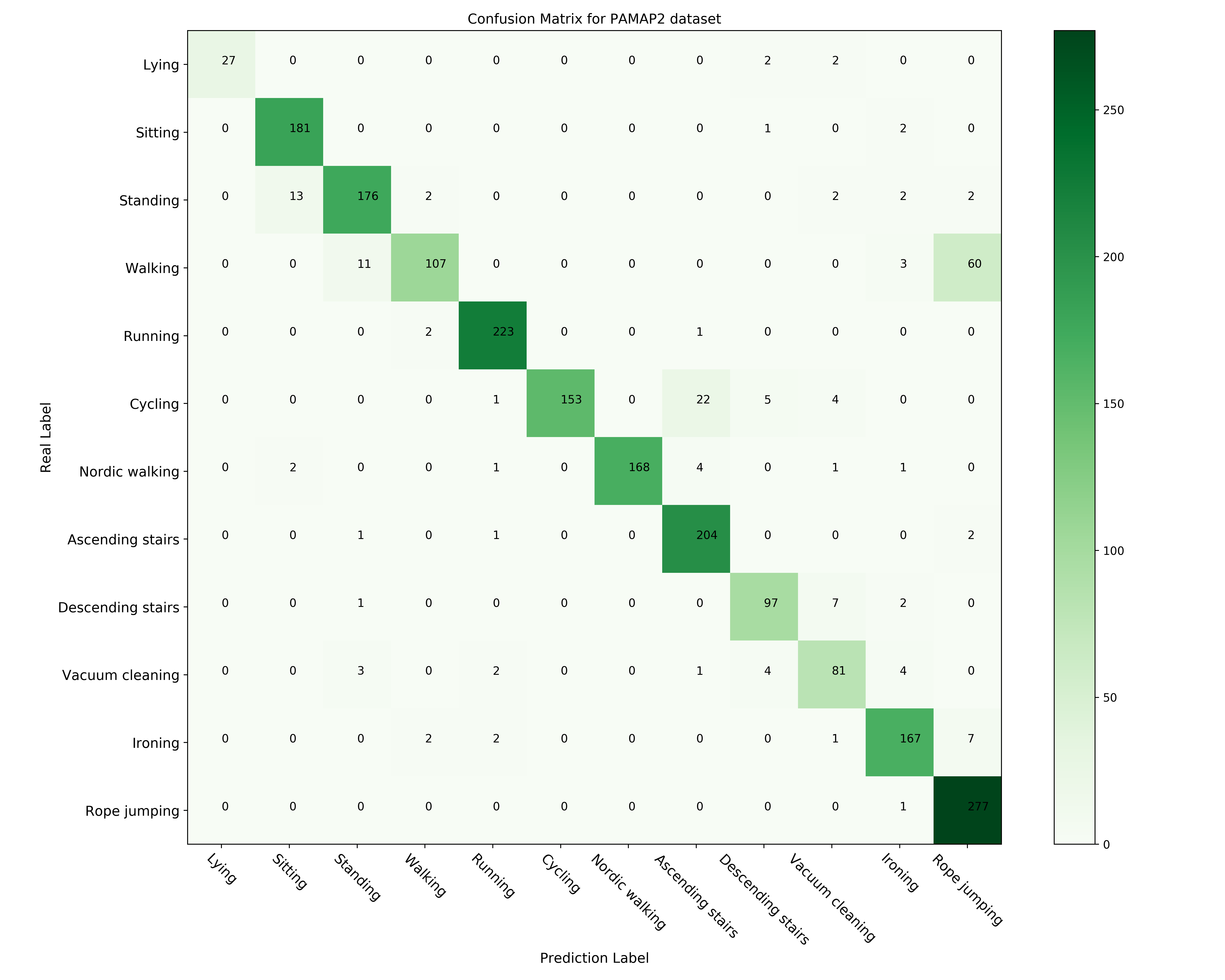}}		 	
		\centering\hspace*{-4.5cm}
	\end{minipage}

	\begin{minipage}{0.5\linewidth}
		{\centering\includegraphics[width=9cm]{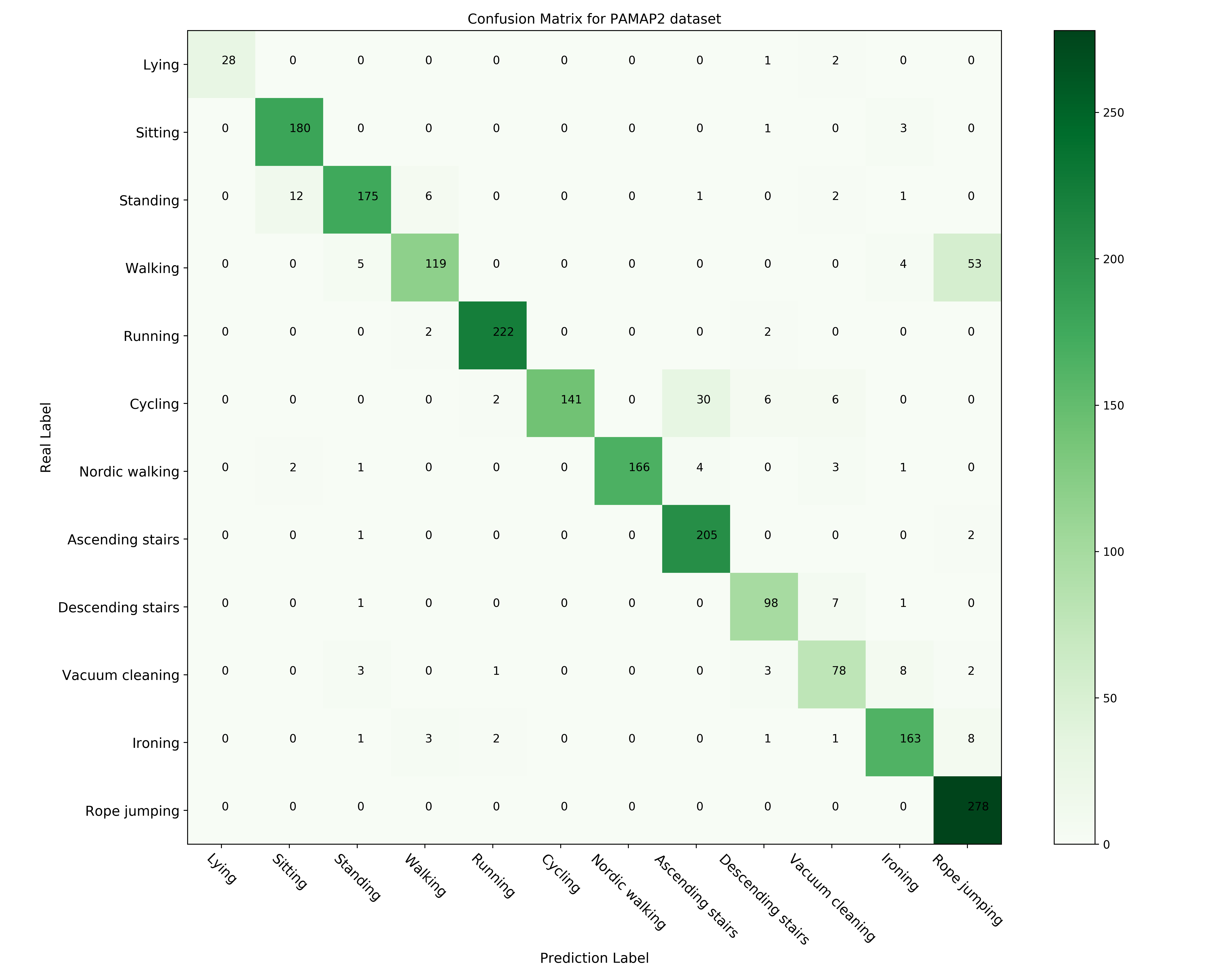}}
		\centering\hspace*{-4.5cm}			
	\end{minipage}
	
	\begin{minipage}{0.5\linewidth}
		{\centering\includegraphics[width=9cm]{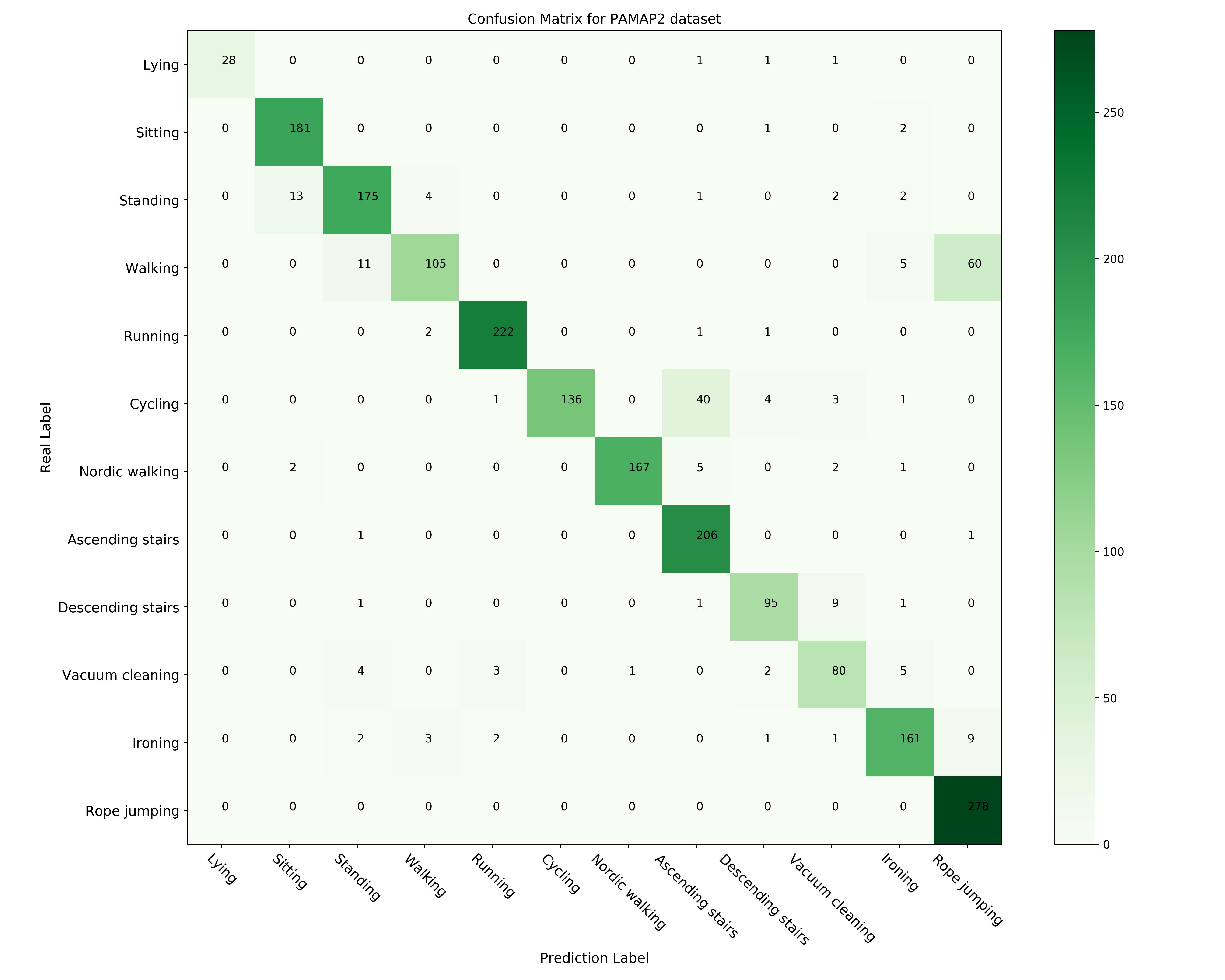}}		 	
		\centering\hspace*{-4.6cm}
	\end{minipage}
	
	\caption{\label{Fig9}Confusion matrix for \textbf{PAMAP2} dataset between the baseline and the Lego CNN. From top to bottom, confusion matrix for the baseline, Lego CNN(o=2,m=0.5), and Lego CNN(o=4,m=0.25)}
\end{figure}
	\indent {To analyze the results in more detail, we show the confusion matrices for the PAMAP2 dataset using the baseline CNN, Lego CNN(o=2,m=0.5) and Lego CNN(o=4,m=0.25), as can be seen in Fig. 8. The three confusion matrices indicate that many of the misclassification are due to confusion between these activities, e.g., \emph{“Ascending stairs” and “Cycling”, ”Rope jumping” and “Walking}. This is because the signal vibration in these two cases are similar. From the results, it can be observed that the Lego CNN can perform comparably well with the baseline CNN. The confusion matrices show similar outputs varying slightly in the classification accuracy. There is no significant decrease on classification performance as compression ratio increases.}
	\begin{figure}[htbp]
		\hspace*{0cm}
		\centering
		\includegraphics[scale=0.16]{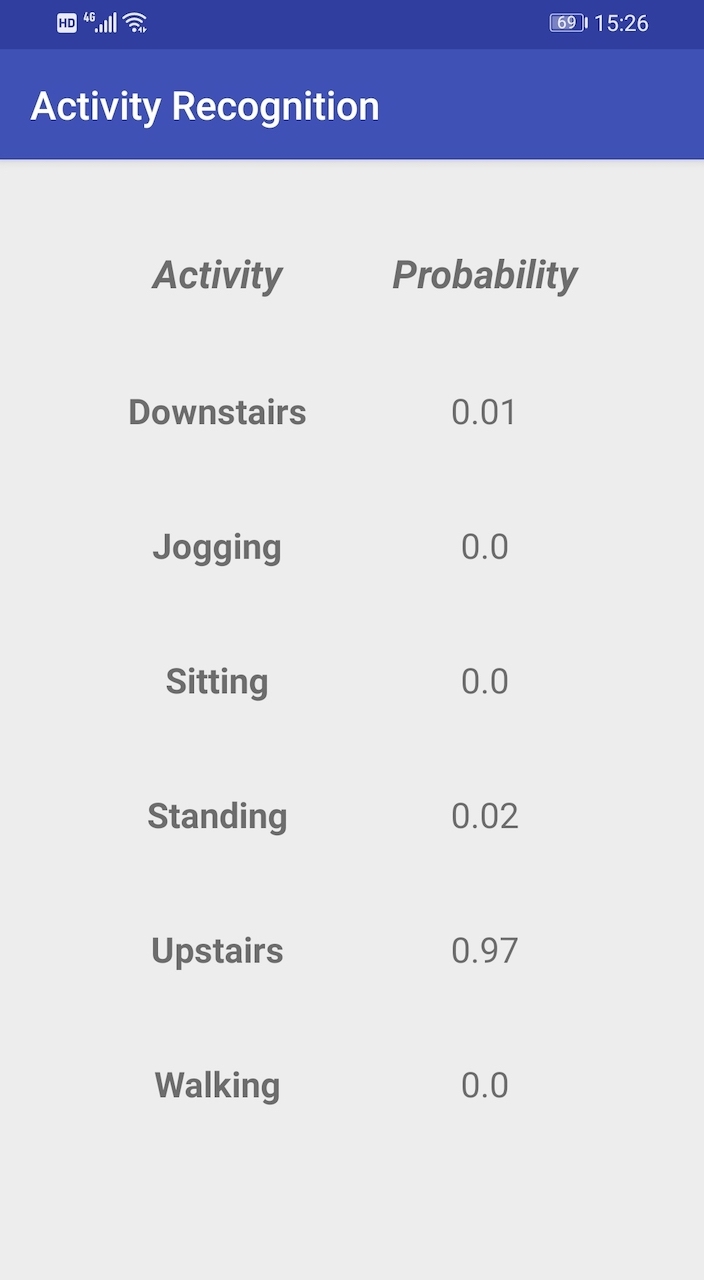}% Here is how to import EPS art
		%\graphicspath{{'G:\360MoveData\Users\Administrator\Desktop\LATEX_LCOAL_ERROR'}}
		%\DeclareGraphicsExtensions{Fig1.eps}
		\caption{\label{Fig9} Screenshot of real implementation with Lego CNN model}
	\end{figure}
	\\
	\indent {\color{black}Finally, in order to evaluate the performance improvement for the practical implementation, we test the proposed deep learning algorithms on an Android smartphone. As smartphones are more convenient and easier to use, they have been utilized in various HAR tasks, which can be seen as a particular case of modern wearable devices.} The HAR APP system presented in \cite{singh2017human} was used as a reference point for the evaluation, which is a smartphone-based application for mobile activity recognition. A screenshot of the app’s main window is shown in Fig 9. Our experiment was implemented on a Huawei Honor 20i device with the Android OS(10.0.0). Several PyTorch trained models with different compression ratio are used on the WISDM dataset and then deployed to build an Android application that can perform on-device activity recognition. The model is converted into pt file and the PyTorch Mobile is added as a Gradle dependency(Java). The classifications can be performed by loading the saved model with PyTorch Mobile. As shown in Table VIII, due to memory access and other overheads, it can conclude to that the Lego CNN(o=4,m=0.25) with 4x theoretical complexity reduction usually results in a 1.7x actual speedup in the implementation.	
	\\
	\vspace{0cm} 
	\begin{table}[h]
		\caption{Model Inference time for different compression}
		\centering
		\begin{tabular}{cccccc}
			\toprule 
			\textbf{Model}&\textbf{Inference Time}(ms/window)\\
			\midrule
			CNN(Baseline)&146-200ms\\
			\midrule
			Lego CNN(o=2,m=0.5)&110-153ms\\
			Lego CNN(o=4,m=0.25)&85-117ms\\
			\bottomrule
			\label{Tab7}
		\end{tabular}
	\end{table}	
	\section{Conclusion}
	\indent Recently, deep CNNs have achieved state-of-the-art performance on various HAR benchmark datasets, which require enormous resources and is not available for mobile and wearable based HAR. Although a series of lightweight structure designs have demonstrated their success in reducing the computational complexity of CNN on visual tasks. They often rely on special network structures, which have been seldom directly adopted for HAR. On the other hand, less complex models such as shallow machine learning techniques could not achieve good performance. Therefore, it is necessary to develop lightweight deep CNNs to perform HAR. In the paper, we for the first time proposed a lightweight CNN using Lego filters for mobile and wearable based HAR tasks. The conventional filters could be replaced with a set of smaller Lego filter, which never rely on special network structures. The STE method is used to optimize the permutation of Lego filters for a filter module. The three-stage split-transform-merge strategy is utilized to further accelerate intermediate convolutions. {\color{black}Our main contribution is to propose a lightweight HAR with smaller Lego filters. The Lego idea can greatly reduce memory and computation cost over conventional CNN, which is accompanied by a slight decrease on performance. To alleviate this, the local loss is used to train the Lego CNN, which can boost the performance without any extra cost. Actually, the local loss may be adding a regularizing effect, which encourages examples from distinct classes to have distinct representations, measured by the cosine similarity. This also can be seen as a kind of supervised clustering \cite{teng2020layer}\cite{nokland2019training}.}\\
	\indent {\color{black}In the paper, the proposed method is tested with smartphones, as well as multiple sensor nodes. To make fair comparison, we evaluate the performance of the proposed method on five public benchmark HAR datasets, which can be classified as both case: the UCI-HAR, UNIMIB-SHAR and WISDM dataset are collected from smartphones; the OPPORTUNITY and PAMAP2 dataset are collected from multiple sensor nodes. In comparison with deployment of multiple sensors nodes, smartphones are cheaper and easier to use, which can be seen as a particular case of wearable devices. Our research mainly focuses on lightweight deep learning implementation of mobile and wearable based HAR.} In order to get better insights on the actual system performance, we evaluate the model performance by using confusion matrix to associate explicit feature representation. On the whole, the results on multiple benchmark datasets suggests that the proposed Lego CNN with local loss is smaller, faster and more accurate. \\
	\\
	\\

	\vspace{-1cm} 
\bibliographystyle{IEEEtran}
\bibliography{ref}

	\vspace{-1.35cm}

	\end{document}